\documentclass[10pt,twocolumn,letterpaper]{article}

\usepackage{config/cvpr}
\usepackage[accsupp]{axessibility}
\newcommand{\moniker}{\mbox{\nameCOLOR{InterDyn}}\xspace}
\definecolor{cvprblue}{rgb}{0.21,0.49,0.74}
\usepackage[pagebackref,breaklinks,colorlinks,allcolors=cvprblue]{hyperref}

\usepackage{fontawesome}
\usepackage{lipsum}
\usepackage[normalem]{ulem}
\usepackage{ifthen}
\usepackage{xstring}
\usepackage{xcolor, colortbl}
\usepackage{multirow}
\usepackage{subcaption}
\usepackage{booktabs}
\usepackage[symbol]{footmisc}
\usepackage{stfloats}
\usepackage{balance}
\usepackage{float}
\usepackage{pifont}

\newcommand{\qheading}[1]{\noindent\textbf{#1.}}

\newcommand{\nameCOLOR}[1]{#1}

\newcommand{\cmark}{\color{green}\ding{51}}
\newcommand{\xmark}{\color{red}\ding{55}}

\newcommand{\NA}{---}

\renewcommand{\etal}{\mbox{et al.}\xspace}
\renewcommand{\ie}{\mbox{i.e.}\xspace}
\renewcommand{\eg}{\mbox{e.g.}\xspace}

\definecolor{turquoise}{cmyk}{0.65,0,0.1,0.3}
\definecolor{purple}{rgb}{0.65,0,0.65}
\definecolor{dark_green}{rgb}{0, 0.5, 0}
\definecolor{orange}{rgb}{0.8, 0.6, 0.2}
\definecolor{darkred}{rgb}{0.6, 0.1, 0.05}
\definecolor{blueish}{rgb}{0.0, 0.3, .6}
\definecolor{light_gray}{rgb}{0.7, 0.7, .7}
\definecolor{pink}{rgb}{1, 0, 1}
\definecolor{greyblue}{rgb}{0.25, 0.25, 1}
\definecolor{forestgreen}{rgb}{0.0, 0.2, 0.13}
\definecolor{darkolivegreen}{rgb}{0.33, 0.42, 0.18}
\definecolor{teaserred}{rgb}{1, 0.588, 0.553}
\definecolor{teaserblue}{rgb}{0.337, 0.757, 1}

\newcommand{\HOI}{\mbox{HOI}\xspace}

\newcommand{\smplX}{\smplX}
\newcommand{\MANO}{\mbox{\nameCOLOR{MANO}}\xspace}

\newcommand{\TI}{\mbox{T2V}\xspace}

\newcommand{\CNN}{\mbox{CNN}\xspace}
\newcommand{\SVD}{\mbox{SVD}\xspace}

\newcommand{\UNET}{\mbox{U-Net}\xspace}
\newcommand{\CONTROLNET}{\mbox{ControlNet}\xspace}
\newcommand{\ControlNet}{\CONTROLNET}
\newcommand{\CLIP}{\mbox{CLIP}\xspace}

\newcommand{\DEXYCB}{\mbox{\nameCOLOR{DexYCB}}\xspace}
\newcommand{\DexYCB}{\DEXYCB}
\newcommand{\SOMETHING}{\mbox{\nameCOLOR{Something-Something-v2}}\xspace}
\newcommand{\SSV}{\mbox{\nameCOLOR{SSV2}}\xspace}

\newcommand{\CLEVRER}{\mbox{\nameCOLOR{CLEVRER}}\xspace}
\newcommand{\SOMETHINGELSE}{\mbox{\nameCOLOR{Something-Else}}\xspace}

\newcommand{\CosHand}{\mbox{\nameCOLOR{CosHand}}\xspace}
\newcommand{\seer}{\nameCOLOR{\mbox{Seer}}\xspace}
\newcommand{\dynamicrafter}{\nameCOLOR{\mbox{DynamiCrafter}}\xspace}
\newcommand{\SAM}{\mbox{\nameCOLOR{SAM2}}\xspace}
\newcommand{\OPENPOSE}{\mbox{\nameCOLOR{OpenPose}}\xspace}
\newcommand{\GDINO}{\mbox{\nameCOLOR{GroundingDINO}}\xspace}
\newcommand{\COTRACKER}{\mbox{\nameCOLOR{CoTracker3}}\xspace}

\newcommand{\LORA}{\mbox{LoRA}\xspace}
\newcommand{\SD}{\mbox{Stable Diffusion}\xspace}
\newcommand{\sd}{\mbox{SD}\xspace}

\newcommand{\MCVD}{\nameCOLOR{\mbox{MCVD}}\xspace}
\newcommand{\UCG}{\nameCOLOR{\mbox{UCG}}\xspace}
\newcommand{\IPP}{\nameCOLOR{\mbox{IPix2Pix}}\xspace}
\newcommand{\TCG}{\nameCOLOR{\mbox{TCG}}\xspace}

\newcommand{\EDM}{\mbox{EDM}\xspace}

\newcommand{\FID}{\mbox{FID}\xspace}
\newcommand{\KID}{\mbox{KID}\xspace}

\newcommand{\FVD}{\mbox{FVD}\xspace}
\newcommand{\KVD}{\mbox{KVD}\xspace}

\newcommand{\LPIPS}{\mbox{LPIPS}\xspace}
\newcommand{\PSNR}{\mbox{PSNR}\xspace}
\newcommand{\SSIM}{\mbox{SSIM}\xspace}
\newcommand{\MF}{\mbox{Motion Fidelity}\xspace}

\newcommand{\RGB}{\mbox{RGB}\xspace}
\newcommand{\FPS}{\mbox{FPS}\xspace}
\newcommand{\MOTIONID}{\mbox{motion ID}\xspace}

\newcommand{\TwoD}{\mbox{2D}\xspace}
\newcommand{\ThreeD}{\mbox{3D}\xspace}

\newcommand{\squareres}{\mbox{256$\times$256}\xspace}
\newcommand{\videores}{\mbox{256$\times$384}\xspace}

\newcommand{\gN}{\mathcal{N}}

\newcommand{\noise}{\boldsymbol{\epsilon}}

\title{\vspace{-0.2 em} \moniker: Controllable Interactive Dynamics with Video Diffusion Models \vspace{-0.2 em}}

\begin{document}

\author{
    \fontsize{10.5}{12}\selectfont
    Rick Akkerman$^{1,2\ast}$
    \quad 
    Haiwen Feng$^{1\ast \dagger}$
    \quad 
    Michael J. Black$^1$ 
    \quad 
    Dimitrios Tzionas$^2$
    \quad 
    Victoria Fern{\'{a}}ndez Abrevaya$^1$ \\
    {\small
    $^1$Max Planck Institute for Intelligent Systems, T{\"u}bingen, Germany
    \quad
    $^2$University of Amsterdam, the Netherlands
}\\
{
    \tt
    \small
    \{rick.akkerman, haiwen.feng, black, victoria.abrevaya\}@tuebingen.mpg.de \quad d.tzionas@uva.nl}
}

\twocolumn[{%
\renewcommand\twocolumn[1][]{#1}%
\maketitle
\begin{center}
    \vspace{-1.4em}
    \centering
    \captionsetup{type=figure}
    \includegraphics[width=\textwidth]{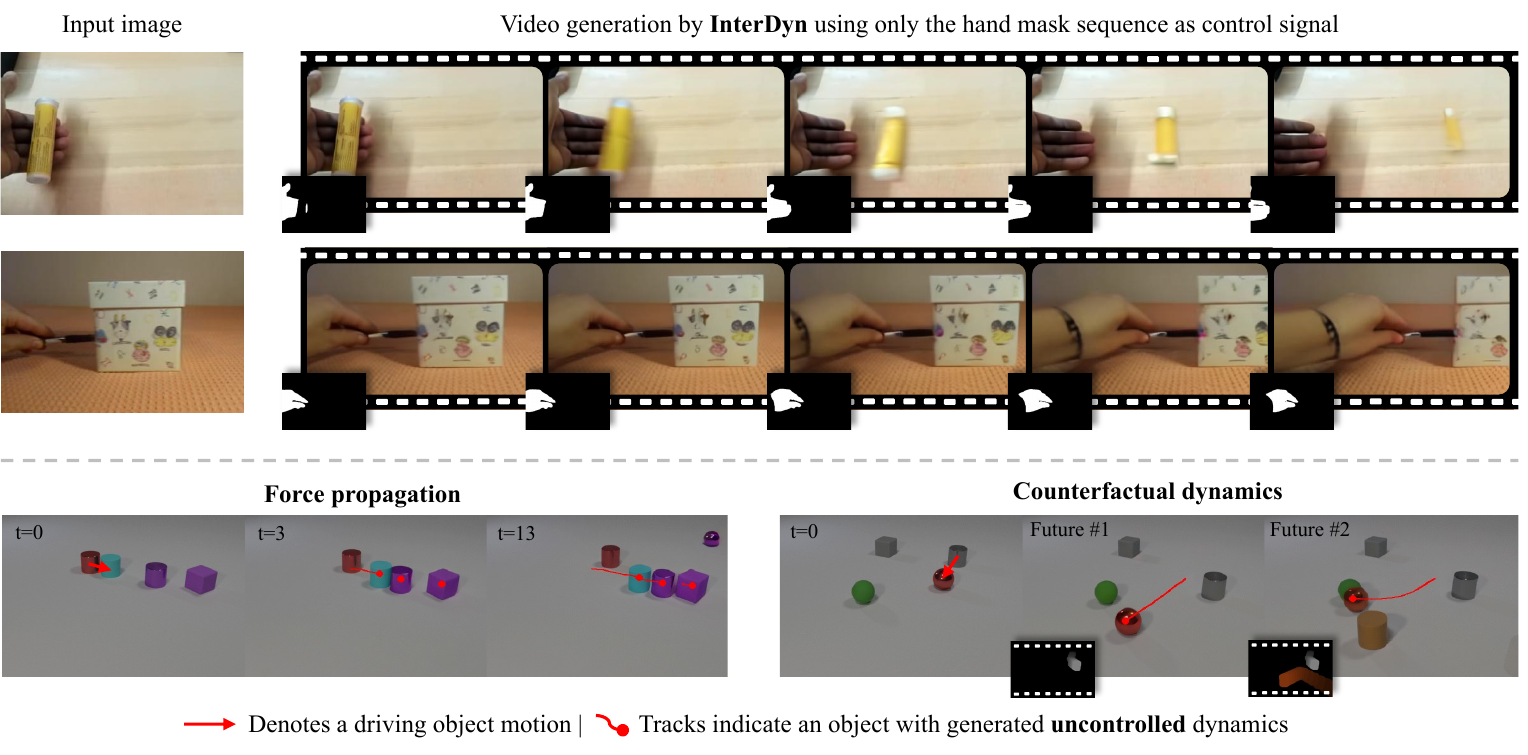}
    \caption{
        We present \textbf{\moniker}, a framework for synthesizing realistic interactive dynamics without \ThreeD reconstruction and physics simulation. Our core principle is to rely on the implicit physics knowledge embedded in large-scale video generative models. Given an image and a ``driving motion'', our model generates the consequential scene dynamics. We investigate the generated interactive dynamics in a simple object collision scenario (bottom) and complex in-the-wild human-object interaction (top).
        } \label{fig:teaser}
    \vspace{+1.6 em}
\end{center}
}]
\maketitle

\def\thefootnote{*}\footnotetext{Equal contribution}
\def\thefootnote{$\dagger$}\footnotetext{Project lead}

\begin{abstract}
Predicting the dynamics of interacting objects is essential for both humans and intelligent systems. However, existing approaches are limited to simplified, toy settings and lack generalizability to complex, real-world environments. Recent advances in generative models have enabled the prediction of state transitions based on interventions, but focus on generating a single future state which neglects the continuous dynamics resulting from the interaction. To address this gap, we propose \moniker{}, a novel framework that generates videos of interactive dynamics given an initial frame and a control signal encoding the motion of a driving object or actor. Our key insight is that large video generation models can act as both neural renderers and implicit physics ``simulators'', having learned interactive dynamics from large-scale video data. To effectively harness this capability, we introduce an interactive control mechanism that conditions the video generation process on the motion of the driving entity. Qualitative results demonstrate that \moniker{} generates plausible, temporally consistent videos of complex object interactions while generalizing to unseen objects. Quantitative evaluations show that \moniker{} outperforms baselines that focus on static state transitions. This work highlights the potential of leveraging video generative models as implicit physics engines. Project page: \url{https://interdyn.is.tue.mpg.de/}.
\end{abstract}
 \newpage
\section{Introduction}
\label{sec:intro}

Humans have the remarkable ability to predict the future dynamics of observed systems intuitively. With just a single image, we can anticipate and imagine how objects will move over time -- not only their motion but also their interactions with the environment and other elements in the scene. Inferring this requires an advanced form of scene-level reasoning beyond merely recognizing the semantics and geometry of static elements; it involves a deep physical and causal understanding of how each object will interact given the environment, object properties, and forces.

There has been a growing interest in developing machine learning systems that emulate similar levels of dynamic understanding given visual observations, such as images or videos. Early work~\cite{wu2017deanimation} addressed this by first reconstructing a \ThreeD representation from the image, then predicting future states with a physics simulator and finally generating the output video with a rendering engine. This relies heavily on explicit reconstruction and simulation, which is computationally intensive, prone to errors, and may not generalize well. More recent methods~\cite{kipf2018neural, baradel2020cophy,
Guen2020DisentanglingPD, li2020causal, janny2022filteredcophy} leverage keypoint or latent representations within graph relational frameworks; however, they have only been trained and validated in over-simplified, synthetic environments, showing limited generalizability to complex real-world scenarios.

\begin{figure}
    \centering
    \includegraphics[width=1.0\columnwidth]{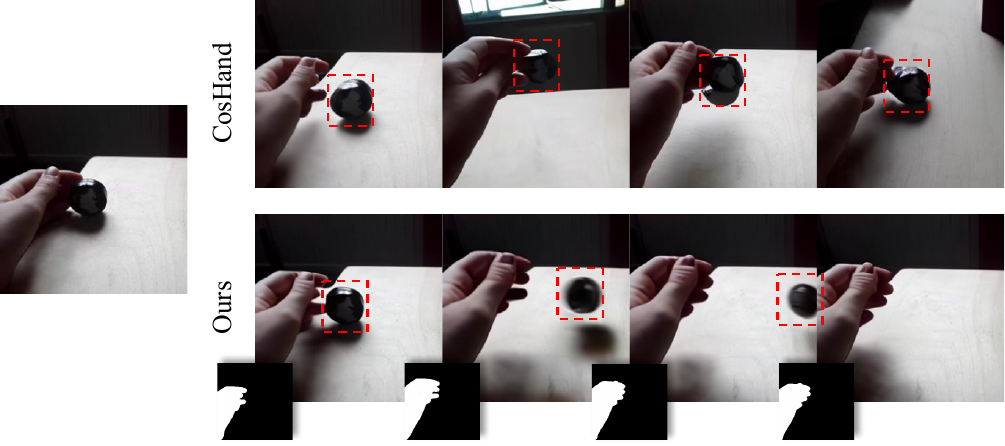}
    
    \caption{
        \qheading{State transition vs.\ dynamics}
        Methods that generate static state transitions (\ie predict a future image) such as \CosHand~\cite{sudhakar2024coshand} struggle to capture the inherent dynamic processes involved in human-object interactions. Here, we show a video sequence where the motion continues beyond the interaction.
    } \label{fig:state-vs-dynamics}
    \vspace{-1.0 em}
\end{figure}

Instead, the advent of powerful generative models~\cite{blattmann2023svd, esser2024scaling, liu2024sora, rombach2022high, bar2024lumiere} opens new avenues for synthesizing interactions under complex scenarios. For example, \mbox{Sudhakar}~\etal~\cite{sudhakar2024coshand} recently proposed \CosHand, a controllable image-to-image model based on \SD~\cite{rombach2022high} that infers \textit{state transitions} of an object. The task here is defined as follows: given an image of a hand interacting with an object, alongside a hand mask of the current frame and a mask of the hand at a future frame, generate a modified input image that satisfies the mask, with realistic interactions. The challenge, as in early intuitive physics works, lies in accurately modeling how the objects will change after forces are applied. However, we argue that static state transitions are insufficient for this task, as they fail to capture the continuous dynamic processes inherent to the problem, \eg see \cref{fig:state-vs-dynamics}. Investigating interactive dynamics within a two-state setting is highly limiting, since dynamics can extend beyond the period of direct contact -- for example, predicting the motion occurring while a person pours water requires a physical understanding that goes beyond the state of the hand at a future frame. The driving force, in this case the hand, may interact with the system only briefly, but the system's subsequent dynamics continue according to physical laws and may even influence other parts via force propagation.

In this paper, we explore \emph{controllable synthesis of interactive dynamics}--generating a video from an input image and a dynamic control signal (\eg a moving hand mask) to model realistic object dynamics. In particular, we propose \moniker{}, a novel framework for synthesizing controllable dynamic interactions that leverages the physical and dynamics ``knowledge'' of a large video generation model~\cite{blattmann2023svd}. Unlike prior approaches that rely on explicit physical simulation~\cite{wu2017deanimation} or are constrained to static state transitions~\cite{sudhakar2024coshand}, we leverage video generation models to generate dynamic processes implicitly, see \cref{fig:teaser}. Specifically, we extend Stable Video Diffusion (\SVD)~\cite{blattmann2023svd} with a dynamic control branch and fine-tune it on diverse scenes, enabling synthesis of complex interactions aligned with the control signal.

We start our investigation by fine-tuning \moniker{} on a simple synthetic scenario of cubes, cylinders, and spheres: the \CLEVRER dataset~\cite{yi2020clevrer}. To control the motion we add a mask driving signal that manipulates the movement of some (but not all) of the objects in the scene. We then evaluate how the synthesized trajectories of uncontrolled objects change under various interactions, including multiple objects colliding with each other. This multi-object collision setting allows us to ``probe'' the physical understanding and causal effects of the video diffusion model, and our qualitative experiments show \moniker{}'s ability for counterfactual future prediction and physical force propagation. 

Next, we evaluate how the system performs in a difficult real-world scenario, such as Human-Object Interaction (\HOI). Here, the dexterity of hand motions and the diversity of objects vastly increase the complexity of the problem. We fine-tune the model on a commonly used \HOI video dataset~\cite{goyal2017something} and compare with the state-of-the-art baseline \CosHand~\cite{sudhakar2024coshand}, as well as two text-control based interactive dynamics generation methods: \seer~\cite{gu2023seer} and \dynamicrafter~\cite{xing2024dynamicrafter}. We quantify our investigations using standard image and video metrics, as well as a motion fidelity metric based on point tracking. \moniker{} surpasses the previous SOTA over 37.5\% on \LPIPS and 77\% on \FVD on the \SOMETHING (\SSV) dataset~\cite{goyal2017something}. Our experiments also demonstrate diverse and physically plausible generations of interactive dynamics, probing into \SVD's ``understanding'' of physics and dynamics.

 In summary, we present \moniker{}, a framework that employs video generative models to simulate object dynamics without explicit \ThreeD reconstruction or physical simulation. We demonstrate how the inherent ``knowledge'' within video generation models can be leveraged to predict complex object interactions and movements over time, implicitly modeling physical and causal dynamics. We perform comprehensive experiments on multi-object collision datasets and hand-object manipulation datasets, demonstrating the effectiveness of our approach. 
\section{Related Work}
\label{sec:related}

\qheading{Modeling human-object interactions (\HOI)} 
Human-object interaction has been widely studied within the context of \ThreeD reconstruction~\cite{hasson2019learning, hasson2021towards, tse2022collaborative, ye2023diffusion, fan2023arctic, fan2024hold, xie2024template}, where the goal is to recover realistic geometry of hands and objects. The field of \ThreeD \HOI synthesis has also received increasing attention, including the generation of static~\cite{taheri2020grab, jiang2021hand, zhu2021toward, kim2024beyond} or dynamic~\cite{zhou2024gears, zheng2023cams, taheri2023grip, paschalidis20243d} hand poses conditioned on \ThreeD objects, whole-body interactions~\cite{xu2023interdiff}, or more recently, hand-object meshes given textual descriptions~\cite{wu2024thor, cha2024text2hoi, ye2024ghop, lu2024ugg, diller2024cghoi}. Few works address \HOI synthesis in the \TwoD domain. \mbox{GANHand}~\cite{corona2020ganhand} predicts \ThreeD hand shape and pose given an \RGB image of an object, while \mbox{AffordanceDiffusion}~\cite{ye2023affordance} estimates a \TwoD hand using a diffusion model. \mbox{Kulal}~\etal~\cite{kulal2023putting} take as input an image of a human and a scene separately and generate a composite image that positions the human with correct affordances. Also relevant is \mbox{HOIDiffusion}~\cite{zhang2024hoidiffusion}, in which a texture-less rendering of a \ThreeD hand and object is converted to a realistic image using a text description. Most closely related to us is \CosHand~\cite{sudhakar2024coshand}, which takes as input an \RGB image of a hand-object interaction, a hand mask at the current state, and the hand mask of the future state, and generates an \RGB image of the future state. Unlike us, they cannot generate post-interaction object dynamics. Importantly, none of these works study \textit{dynamics}, generating instead discrete state transitions that fail to capture the nuanced, temporally coherent behaviors observed in interactions.

\vspace{0.12cm}
\qheading{Synthesizing causal physical relations from visual input}
A growing body of work aims to model and predict physical causal effects from visual inputs such as images or videos. For example, research in intuitive physics seeks to replicate the human-like, non-mathematical understanding of physical events, \eg by predicting future frames given an input video. Early works like~\cite{lerer2016learning, groth2018shapestacks} train neural networks to assess the stability of block towers, while \cite{Guen2020DisentanglingPD} leverage prior physical knowledge formalized through partial differential equations (PDEs). Other approaches investigate counterfactual reasoning by leveraging graph neural networks~\cite{baradel2020cophy, janny2022filteredcophy}. \mbox{Wu}~\etal~\cite{wu2015galileo, wu2016physics, wu2017deanimation} explore the use of an inverse rendering approach, extracting geometry and physical properties from the video which are then coupled with a physics simulator and a rendering engine to generate the future frames. Other works~\cite{watters2017visual} incorporate \mbox{Interaction Networks}~\cite{battaglia2016interaction} to approximate physical systems from video data. These approaches are often limited to simplified, synthetic datasets and struggle to generalize to real-world scenarios.

Recent methods have started to combine language models with physical engines. \mbox{Liu}~\etal~\cite{liu2023mind} ground a large language model using a computational physics engine while \mbox{Gao}~\etal~\cite{gao2024physically} show that fine-tuning a vision-language model (VLM) on annotated datasets of physical concepts improves its understanding of physical interactions. Closely related to our work is \mbox{PhysGen}~\cite{liu2024physgen}, which trains an image-to-video model that conditions the video generation on physics parameters (\eg, force or torque). However, the model relies on a dynamics simulator to generate motion, and its application is limited to rigid objects. A related but tangential line of work focuses on identifying and generating the effects of objects on their surroundings. For example, \mbox{Omnimatte}~\cite{lu2021omnimatte} introduces the problem of identifying all parts of a scene influenced by an object, given a video and a mask of the object. Similarly, \mbox{Lu}~\etal~\cite{lu2020layered} propose to re-time the motion of different subjects in a scene while maintaining realistic interactions with the environment. \mbox{ActAnywhere}~\cite{pan2024actanywhere} generates videos with plausible human-scene interactions, taking a masked video of a person and a background image as input. These works address the problem of synthesizing realistic interactions within a scene, however, lack fine-grained control.

\qheading{Controllable video generation}
Video generation has advanced significantly in recent years, with diffusion models leading to substantial improvements in unconditional~\cite{ho2022video, yu2023video}, text-based~\cite{singer2022make, ho2022imagen, blattmann2023svd, hu2022make, chen2023videocrafter1, zhou2022magicvideo, blattmann2023align, gu2023seer, bar2024lumiere, girdhar2024emu, yang2024cogvideox, wang2023modelscope, guo2023animatediff, wu2023tuneavideo, xing2024dynamicrafter} and image-based~\cite{blattmann2023svd, xing2024dynamicrafter, wang2024videocomposer, bar2024lumiere, girdhar2024emu, guo2024i2vadapter} generation. These advances have raised the question of how to incorporate more nuanced control into video generation. Some text-to-video approaches are trained by ``inflating'' text-to-image (\TI) models~\cite{wu2023tuneavideo, hu2023videocontrolnet, chen2023controlavideo, guo2024i2vadapter, cai2024generative, guo2023animatediff}, and can thus be integrated with conditional \TI models such as \ControlNet~\cite{zhang2023adding} or \TI-Adapter~\cite{mou2024t2iadapter}. Control can also be achieved by conditioning on trajectories~\cite{wu2024draganything, ma2023trailblazer, yin2023dragnuwa} or bounding-boxes~\cite{wang2024boximator}, fine-tuning on appropriate datasets. \mbox{VideoComposer}~\cite{wang2024videocomposer} incorporates multiple condition types, including text, depth, style, and temporal conditions via motion vectors. Camera motion control has also been explored, with \mbox{AnimateDiff}~\cite{wang2004image} employing \LORA~\cite{hu2022lora} modules to control camera movement, while \mbox{MotionCtrl}~\cite{wang2024motionctrl} and \mbox{CameraCtrl}~\cite{he2024cameractrl} directly embed the camera information for more precise control. Additionally, several works target human animation from a pose control signal, such as \mbox{DreamPose}~\cite{karras2023dreampose}, \mbox{MagicPose}~\cite{xu2024magicanimate}, and \mbox{AnimateAnyone}~\cite{hu2024animateanyone}, but do not account for interactions.
\section{Controllable Interactive Dynamics}
\label{sec:method}

Video diffusion models such as~\cite{blattmann2023svd, liu2024sora} have demonstrated impressive performance in generating videos from text or images, and have even shown potential in tasks that require \ThreeD understanding when properly fine-tuned~\cite{hu2024depthcrafter, voleti2024sv3d}. Trained on millions of videos, we hypothesize that these models also possess implicit knowledge of complex interactive dynamics, such as those that appear when humans interact with objects. Out of the box, however, they lack a precise control mechanism, often relying solely on textual inputs or requiring careful selection of the starting frame.

\vspace{0.4em}
\qheading{Task}
Given an input image, $\boldsymbol x \in \mathbb{R}^{1 \times H \times W \times 3}$, and a \textit{driving motion} in the form of a pixel-wise corresponding control signal $\boldsymbol c \in \mathbb{R}^{N \times H \times W \times 3}$, we task \moniker{} with generating a video sequence, $\boldsymbol y \in \mathbb{R}^{N \times H \times W \times 3}$, depicting plausible object dynamics. Through this task, we aim to learn the conditional distribution between a driving motion, such as that of a human hand, and the consequent motion of manipulated objects. In other words, the model needs to synthesize plausible object movement and appearance \textit{without any indication} other than the driving motion, while maintaining physical and visual consistency with the input image.

\vspace{0.4em}
\qheading{Stable Video Diffusion}
We extend Stable Video Diffusion~\cite{blattmann2023svd} (\SVD) to enable controllable interactive dynamics and explore the versatility of this model across a range of scenarios. \SVD is a publicly available \UNET-based latent diffusion model~\cite{rombach2022high} that extends \SD 2.1 to video generation by interleaving the network with temporal layers. Given a static input image of a scene, \SVD denoises a sequence of $N$ frames $\boldsymbol y \in \mathbb{R}^{N \times H \times W \times 3}$ to generate a video that follows the initial frame. The input image is fed into the denoising \UNET by concatenating its latent to each of the frames in the noised input, and by supplying its \CLIP~\cite{radford2021learning} embedding to the \UNET's cross-attention layers. In addition, \SVD is conditioned on the video's \FPS and \MOTIONID, where the \MOTIONID represents the amount of motion in the video. We found a \MOTIONID of 40 to align well with our frozen \SVD prior.

\begin{figure}
    \centering
    \includegraphics[width=1.0 \columnwidth]{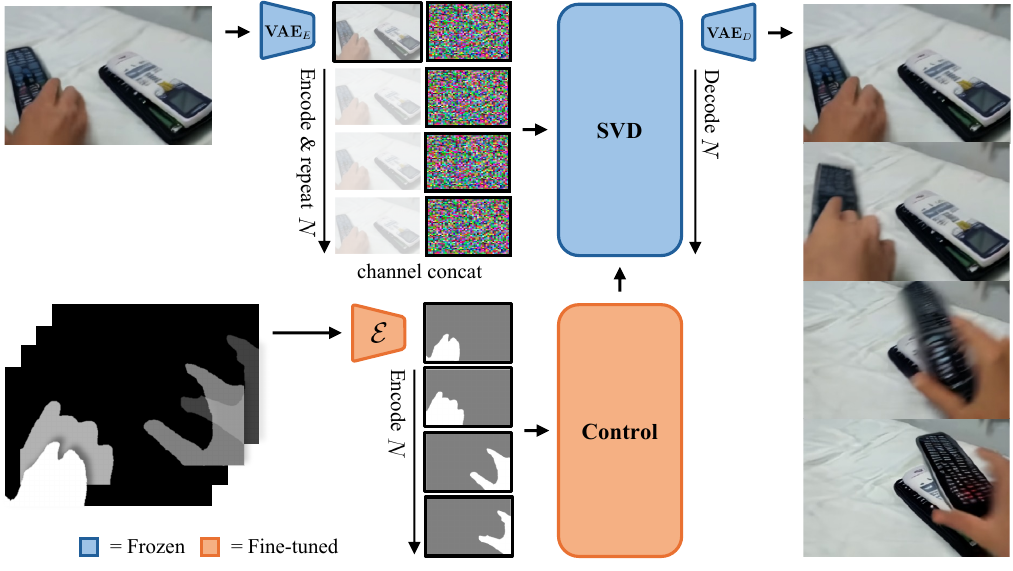}
    \caption{
        \qheading{Overview of \moniker{}}
        Given an input image depicting a scene, such as a hand holding a remote, and a ``driving motion,'' such as a sequence of binary hand masks, \moniker generates a video depicting plausible hand and object dynamics.
        Crucially, \moniker receives no control signal for the object.
        Through this setup, we probe and assess the implicit knowledge of large video generation models on complex interactive dynamics.
        We use Stable Video Diffusion (SVD) as our frozen backbone and fine-tune a separate control signal encoder. Videos are iteratively denoised over $T$ timesteps, starting from Gaussian noise $\noise \sim \gN(0, I)$. 
    } \label{fig:pipeline}
\end{figure}

\vspace{0.4em}
\qheading{Control}
\moniker{} extends \SVD with an additional control signal $\boldsymbol c \in \mathbb{R}^{N \times H \times W \times 3}$ by integrating a \CONTROLNET-like branch~\cite{zhang2023adding}. An overview of our pipeline is presented in \cref{fig:pipeline}. The \SVD weights remain frozen to preserve its learned dynamics prior. Following~\cite{zhang2023adding}, we introduce a trainable copy of the \SVD encoder $E$, connected to \SVD's frozen decoder via skip connections, and modulated by zero-initialized convolutions. We use a small \CNN, $\mathcal{E}(\cdot)$, to encode the control signal $\boldsymbol c$ into the latent space, which is then added to the noisy input latent that is passed to the \CONTROLNET encoder. Similar to \SVD, the control branch interleaves convolutional, spatial, and temporal blocks, enabling \moniker{} to process the control signal in a temporal-aware manner. This helps \moniker{} to be robust to noisy control signals, see \cref{fig:noisy}. We opt for binary masks as conditioning signal due to their accessibility. However, our method can be extended to incorporate diverse types of signals. We find that for hand-object interactions, the type of conditioning signal does not significantly impact performance, see \cref{sec:ablation} and \cref{tab:2_dexycb}.

\vspace{0.4em}
\qheading{Inference}
During inference, we start from an input image, control signal sequence, and randomly sampled Gaussian noise $\noise \sim \gN(0, I)$. Through iteratively applying \moniker over $T$ denoising timesteps, we generate a video $\boldsymbol y$ depicting plausible hand and object dynamics, aligned with the control signal.
\section{Experiments}
\label{sec:experiments}

The primary goal of this work is to synthesize scene-level interactive dynamics by leveraging the implicit physical understanding of a pre-trained video generative model. We begin by probing the model's ability to predict physically plausible outcomes within simulated environments, specifically using the \CLEVRER dataset~\cite{yi2020clevrer}. We test force propagation amongst uncontrolled objects and examine counterfactual future prediction by generating videos for one input image with different control signals. Motivated by promising results, we extend our investigation to complex, real-world hand-object interaction scenarios using the \SOMETHING (\SSV) dataset~\cite{goyal2017something}, conducting comprehensive comparisons with existing baselines that pursue similar objectives. Additionally, we showcase diverse physical examples to demonstrate the capabilities of \moniker{} in generating realistic interactive dynamics.

\begin{figure*}[t]
    \centering
    \begin{subfigure}{1.0\textwidth}
        \centering
        \includegraphics[width=1.0\textwidth]{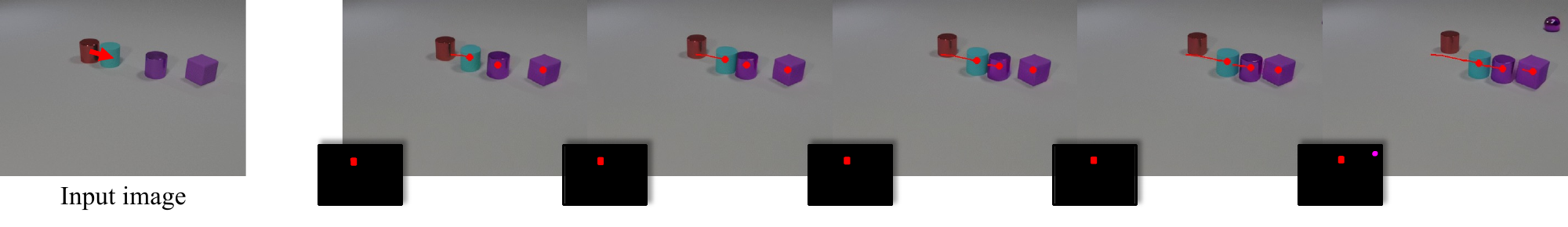}
        \caption{
            \qheading{Force Propagation}
            \moniker{} generates force propagation to and amongst uncontrolled objects (cyan cylinder, pink cylinder, and pink cube).
        } \label{fig:clevrer_force}
        \vspace{0.5em}
    \end{subfigure}
    \begin{subfigure}{1.0\textwidth}
        \centering
        \includegraphics[width=1.0\textwidth]{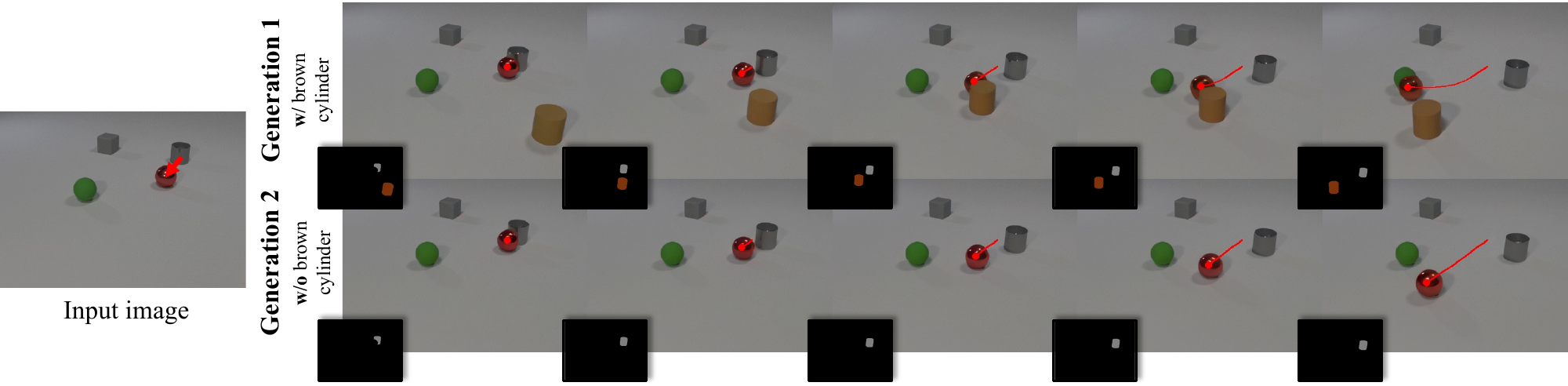}
        
        \vspace{0.5em}
        
        \caption{
            \qheading{Counterfactual Dynamics}
            Two generations of \moniker{} on the same input image: (1) w/ brown-cylinder control and (2) w/o brown-cylinder control.
        } \label{fig:clevrer_future}
    \end{subfigure}
    \caption{
        \qheading{Qualitative investigation on the \CLEVRER dataset}
        Given an input image and the ``driving'' motion of one or two objects, our model predicts the future interactive dynamics of multiple elements in the scene. The driving motion is given in the form of semantic mask sequences. The generated object motions are highlighted with a red-line trajectory. Note that our model can generate videos with force propagation across multiple uncontrolled objects (top) and can generate multiple futures (bottom). \faSearch~\textbf{Zoom in} for details.
    } \label{fig:results_clevrer}
    \vspace{-1em}
\end{figure*}

\subsection{Implementation}
\label{sec:training-details}

We initialize \moniker{} with the 14-frame image-to-video weights of \SVD \cite{blattmann2023svd}. During training, we use the Adam optimizer \cite{kingma2015adam} with a learning rate of \( 1 \times 10^{-5} \). We use the \EDM framework~\cite{karras2022elucidating} with noise distribution $\log \sigma \sim \gN(0.7, 1.6^2)$. We train on two \mbox{80GB H100} GPUs, with a per-GPU batch size of 4. Video length and \FPS define the temporal resolution of dynamics; to balance short- and long-range events we subsample videos to 7 \FPS. To facilitate classifier-free guidance~\cite{ho2022classifier}, we randomly drop the input image with a probability of 5\%. At inference, we apply the Euler scheduler \cite{karras2022elucidating} over 50 denoising timesteps.

\subsection{Metrics}
\label{sec:metrics}

We evaluate \moniker{} on image quality, spatio-temporal similarity, and motion fidelity. Image quality metrics are computed frame-wise. All metrics are reported excluding the first frame, as it serves as input conditioning.

\vspace{0.35em}
\qheading{Image quality}
We report the Structural Similarity Index Measure (\SSIM)~\cite{wang2004image}, Peak Signal-to-Noise Ratio (\PSNR), Learned Perceptual Image Patch Similarity (\LPIPS)~\cite{zhang2018unreasonable}, Fr\'echet Inception Distance (\FID)~\cite{heusel2017gans} and unbiased Kernel Inception Distance (\KID)~\cite{binkowski2018demystifying}.

\vspace{0.35em}
\qheading{Spatio-temporal similarity}
To assess the spatio-temporal perceptual similarity between the ground truth and the generated video distributions, we use the Fr\'echet Video Distance (\FVD) and unbiased Kernel Video Distance (\KVD) proposed in \cite{unterthiner2019fvd}. We use the implementation of \cite{skorokhodov2022stylegan}.

\vspace{0.35em}
\qheading{Motion Fidelity}
Through \moniker{}, we do not have explicit control over object dynamics, which means that the pixel alignment of an object in the generated and ground truth video is only guaranteed in the starting frame. In this case, comparing generated object motion to the ground truth naively might misrepresent the true quality of object motion over time. Therefore, we adapt the \MF (MF) metric proposed by \mbox{Yatim}~\etal~\cite{yatim2024space}, which measures the similarity between point-tracking trajectories.

To compute the metric for any video, we obtain a mask of the object in the starting frame, sample 100 points on the object, and track these throughout both the ground truth and generated video using \COTRACKER~\cite{karaev2024cotracker3}. Given the resulting two sets of tracklets $\mathcal{T}=\{\tau_1,\dots,\tau_n\},\tilde{\mathcal{T}}=\{\tilde{\tau}_1,\dots,\tilde{\tau}_m\}$ the motion fidelity metric is defined as:
\begin{equation}
    \frac{1}{m}\sum_{\tilde{\tau}\in \tilde{\mathcal{T}}} \underset{\tau \in \mathcal{T} }{\text{max}} \ \textbf{corr} (\tau,\tilde{\tau})+\frac{1}{n}\sum_{{\tau}\in {\mathcal{T}}} \underset{\tilde{\tau} \in \tilde{\mathcal{T}} }{\text{max}} \ \textbf{corr} (\tau,\tilde{\tau})\text{,}
  \label{eq:motion-fidelity}
\end{equation}
with the correlation between two tracklets $\textbf{corr}(\tau,\tilde{\tau})$~\cite{liu2005motion}:
\begin{equation}
\textbf{corr}(\tau,\tilde{\tau})
    \frac{1}{F}\sum_{k=1}^F\frac{{v_k^xa \cdot \tilde{v}_k^x + v_k^y \cdot \tilde{v}_k^y}}{{\sqrt{(v_k^x)^2 + (v_k^y)^2} \cdot \sqrt{(\tilde{v}_k^x)^2 + (\tilde{v}_k^y)^2}}}\text{,}
    \label{eq:tracklet_correlation}
\end{equation}
where $(v_k^x,v_k^y),(\tilde{v}_k^x,\tilde{v}_k^y)$ are the $k^{th}$ frame displacement of tracklets $\tau,\tilde{\tau}$ respectively. If there are less than 100 points to query on the object due to it being too small, we do not consider the video for the motion fidelity metric.

\subsection{Probing Dynamics with Object Collision Events}
\label{sec:probing}

Here, we fine-tune \moniker{} on an object collision dataset to \emph{probe} its ability to generate realistic object interactions. Qualitatively, we review whether \moniker{} can produce plausible object motion for uncontrolled objects, given the motion of objects entering the scene. In addition, we examine whether \moniker{} can generate counterfactual videos for the same input image, but different control signals.

\vspace{0.35em}
\qheading{Dataset}
We use \CLEVRER~\cite{yi2020clevrer}, which provides 20,000 videos of colliding objects with annotated segmentation masks and metadata on collision events. We construct a control signal for objects entering the scene and aim to use \moniker{} to generate the motion of the objects that are already present, upon collision. Stationary objects do not receive any form of control signal. Colored masks help the model distinguish unique objects. The frames are cropped and scaled to \(320 \times 448\), and we only sample input frames before collisions between objects in the scene, to maximize \moniker{}'s exposure to interactive dynamics.

\vspace{0.35em}
\qheading{Force propagation}
\moniker{} can generate force propagation dynamics between a controlled object and an uncontrolled object, as well as amongst uncontrolled objects, as illustrated in \cref{fig:clevrer_force}. Here, the red cylinder at the top left is the driving force. It collides with the uncontrolled blue cylinder, which then collides with the uncontrolled purple cylinder, in turn striking the uncontrolled purple cube on the far right. Point-tracking trajectories display how collisions alter each object's position. This suggests that \moniker{} possesses an implicit understanding of physical interactions, enabling it to generate plausible dynamics.

\vspace{0.35em}
\qheading{Counterfactual dynamics}
By altering the control signal, \moniker{} can simulate counterfactual scenarios for the same input image, as shown in \cref{fig:clevrer_future}. In ``Generation 1'', the gray cylinder (controlled) collides with the stationary red sphere (uncontrolled), causing it to move; it is later struck by the brown cylinder (controlled), altering its path once again. In ``Generation 2'', removing the brown cylinder lets the red sphere continue along its original trajectory, consistent with expectations. Crucially, there is no control signal for the red sphere throughout the sequence; its motion and trajectory are entirely generated by \moniker{}.

\vspace{0.35em}

Probing \moniker{} on \CLEVRER highlights its ability to generate interactive dynamics for objects within a simple synthetic environment. We provide additional results in video format on our webpage.

\begin{figure}[b]
    \vspace{-0.5em}
    \centering
    \includegraphics[width=1.0 \columnwidth]{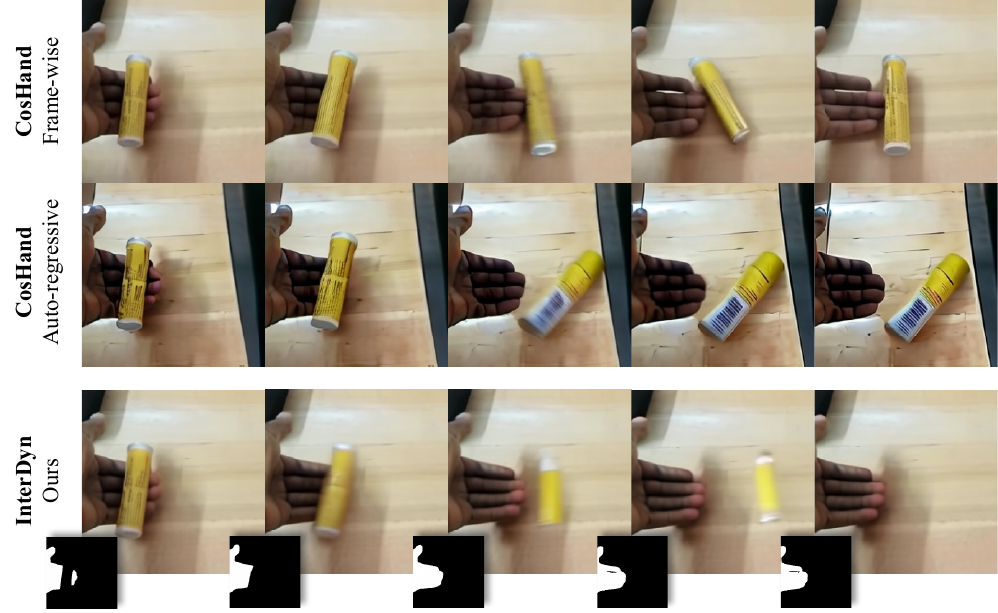}
    \caption{
        \qheading{Qualitative comparison}
        A two-state approach such as \CosHand~\cite{sudhakar2024coshand} struggles with post-interaction object dynamics.
    } \label{fig:baseline_comparison}
\end{figure}
\begin{figure}[b]

    \centering
    \includegraphics[width=\columnwidth]{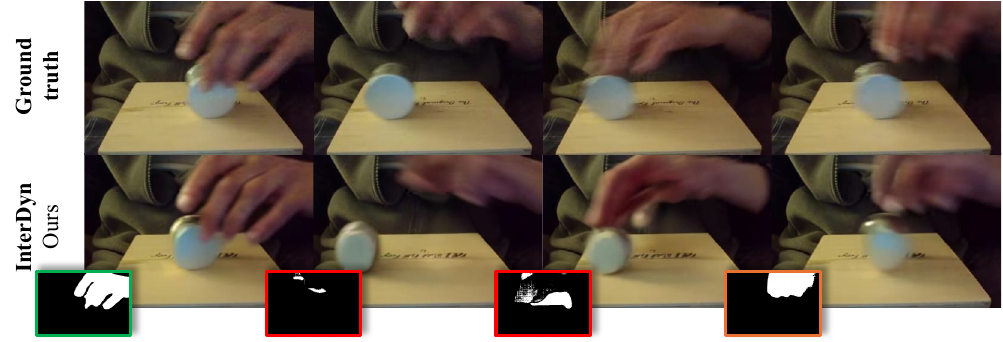}
    
    \caption{
        \qheading{Robustness to noise}
        \SAM outputs noisy/coarse masks for frames with considerable motion blur (orange/red). Despite this, \moniker{} can generate plausible hand and object dynamics.
    } \label{fig:noisy}

    \vspace{-0.5em}
    
\end{figure}
\begin{figure*}
    \centering
    \includegraphics[width=1.0\textwidth]{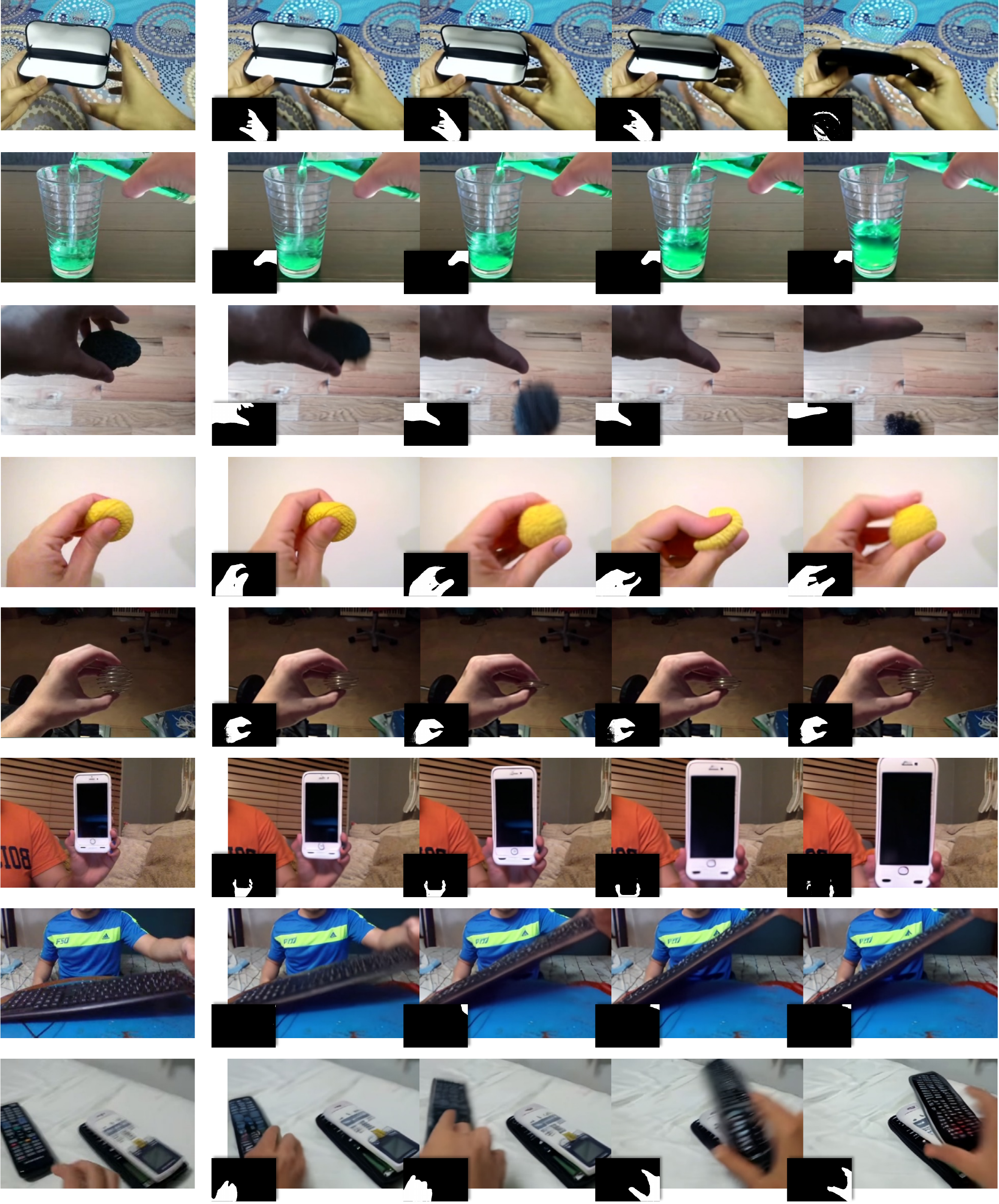}
    \caption{
        \qheading{Diverse generation of interactive dynamics}
        We show multiple challenging examples, such as (from top to bottom): interacting with articulated objects, pouring liquid, letting an object fall, squeezing a highly deformable or ``collapsible'' object, interacting with reflective objects, tilting a ridged object, or stacking objects. \faSearch~\textbf{Zoom in} for details.
    } \label{fig:results_diverse}
\end{figure*}

\subsection{Generating Human-Object Interactions}
\label{sec:generating-hoi}

For this experiment, we fine-tune \moniker{} on a human-action dataset, focused on hand-object interaction. We encode human movement as a sequence of pixel-aligned binary hand masks and task \moniker{} to generate a video with hand movements and corresponding object dynamics.

\vspace{0.25em}
\qheading{Dataset}
We use \SOMETHING~(\SSV), which provides 220,847 videos of humans performing basic actions with everyday objects. It contains actions such as ``pushing \mbox{[something]} from left to right'', ``squeezing \mbox{[something]}'' and ``lifting \mbox{[something]} with \mbox{[something]} on it''. This dataset allows us to train \moniker{} at a larger scale and compare with our baseline \CosHand~\cite{sudhakar2024coshand}. We train one version of \moniker{} at the same resolution as \CosHand, \squareres, and a second version at \videores, which aligns better with the dynamic prior of \SVD.

We generate binary hand masks by prompting Segment Anything 2 (\SAM)~\cite{ravi2024sam} with hand bounding boxes, provided by the \SOMETHINGELSE dataset~\cite{materzynska2020something}. Similar to our baseline \CosHand, we exclude the ``pretending'' class. Additionally, we remove all ``\mbox{[something]} is done behind \mbox{[something]}'' classes since the object would be out of view of the camera. We remove videos smaller than the target resolution/length, ensure the hand and objects are continuously visible, and crop videos larger than the target resolution while centering the object. We include videos without obvious motion and state transition, such as ``holding \mbox{[something]}''. Since bounding box annotations are only provided for the train split (79,043 samples for \videores and 104,260 for \squareres) and validation split (8,667 samples for \videores and 11,229 for \squareres), we report all evaluation metrics on the validation split.

\begin{table*}[t]
    \centering
    \resizebox{\linewidth}{!}{
        \begin{tabular}{@{}lccccccccccccccc@{}}

\toprule

Method &
\SSIM $\uparrow$ & &
\PSNR $\uparrow$ & &
\LPIPS $\downarrow$ & &
\FID$\downarrow$ & &
\KID$\downarrow$ & &
\FVD$\downarrow$ & &
\KVD$\downarrow$ & &
\MF $\uparrow$~\cite{yatim2024space} \\

\midrule

\seer~\cite{gu2023seer} &
0.418 & &
10.71 & &
0.588 & &
33.35 & &
0.030 & &
287.46 & &
81.31 & &
\NA \\

\dynamicrafter~\cite{xing2024dynamicrafter}$^\dagger$ &
\NA & &
\NA & &
\NA & &
17.48 & &
0.014 & &
204.11 & &
31.81 & &
\NA \\

\mbox{\CosHand-Independent}~\cite{sudhakar2024coshand} &
0.615 & &
16.87 & &
0.313 & &
\textbf{2.95} & &
\textbf{0.002} & &
91.18 & &
19.24 & &
0.432 \\

\mbox{\CosHand-Autoregressive}~\cite{sudhakar2024coshand} &
0.531 & &
14.92 & &
0.408 & &
12.66 & &
0.012 & &
90.30 & &
13.68 & &
0.570 \\

\midrule

Ours \squareres &
\underline{0.664} & &
\underline{18.60} & &
\underline{0.260} & &
\underline{4.95} & &
\underline{0.004} & &
\textbf{19.27} & &
\textbf{1.99} & &
\underline{0.633} \\

Ours \videores &
\textbf{0.680} & &
\textbf{19.04} & &
\textbf{0.252} & &
5.07 & &
\underline{0.004} & &
\underline{22.22} & &
\underline{2.09} & &
\textbf{0.641} \\

\bottomrule
        
\end{tabular}
    }
    \vspace{0.3em}
    \caption{
        \qheading{Quantitative comparison on \SOMETHING}
        We compare against two language-instructed video generation methods, \seer~\cite{gu2023seer} and \dynamicrafter~\cite{xing2024dynamicrafter} and two video extensions of our baseline \CosHand~\cite{sudhakar2024coshand}. We report results for \moniker{} at two resolutions: \squareres (matching \CosHand) and \videores (matching \SVD's prior). Methods denoted with $\dagger$ do not use \SSV as their training dataset.
    } \label{tab:results-ssv2-video}
    \vspace{0.2em}
\end{table*}

\vspace{0.25em}
\qheading{Baselines}
To generate videos using \CosHand~\cite{sudhakar2024coshand} rather than state transitions, we run \CosHand in a frame-by-frame approach (\CosHand-Independent) and an auto-regressive approach (\CosHand-Autoregressive). In the frame-by-frame variant, each future frame is independently predicted from the initial frame and its corresponding mask:
\[
\hat{x}_{t+1} = \text{\CosHand}(x_0, h_0, h_{t+1})\text{,} \quad \forall t \in [0, 13],
\]
where \mbox{\( h_0 \)} and \mbox{\( h_{t+1} \)} denote masks, and \mbox{\( x_0 \)} the initial frame.
In the auto-regressive variant, we use \CosHand{} to generate video frames sequentially:
\[
\hat{x}_{t+1} = \text{\CosHand}(\hat{x}_{t}, h_{t}, h_{t+1})\text{,} \quad \forall t \in [0, 13],
\]
with \mbox{\( \hat{x}_{t} \)} being the previously generated frame and \mbox{\( \hat{x}_{0}=x_0 \)}.

We also compare against two language-instructed video generation methods: \seer \cite{gu2023seer} and \dynamicrafter \cite{xing2024dynamicrafter}, which we prompt with a first frame and its corresponding \SSV class label as instruction. Note that \dynamicrafter was trained with a random video frame as conditioning, \ie not always the first frame. At inference, its generated videos are thus not strictly a continuation of the conditioning, which precludes frame-aligned metrics. We compare \moniker{} against these baselines in \cref{tab:results-ssv2-video}.

\vspace{0.25em}
\qheading{Quantitative analysis}
The frame-by-frame variant of \CosHand achieves high image quality but struggles with temporal coherence and motion fidelity. Qualitatively, we notice that object locations are inconsistent across frames. In contrast, the auto-regressive variant improves object motion fidelity but suffers from lower frame-wise image quality due to error propagation. Both variants fail in scenarios requiring accurate post-interaction dynamics, such as when objects continue moving after being released from direct hand contact, as shown in \cref{fig:baseline_comparison}.

Our method, \moniker{} (\videores), achieves the best overall performance, surpassing our baseline \CosHand in spatio-temporal dynamics, motion fidelity, and all but two image quality metrics. We hypothesize that this might be due to two reasons (1) we use a frozen \UNET, while \CosHand fine-tunes its model on \SSV, so \CosHand might generate frames closer to the \SSV distribution, and 2) when \SVD was trained, it was initialized with \sd weights as spatial layers, and then fine-tuned over multiple stages; this might have degraded its spatial prior, and by extension the quality of produced images compared to \CosHand.

Fine-tuned on noisy masks and leveraging its temporal-aware control branch, \moniker{} can interpret a noisy control sequence; \eg when \SAM produces a coarse and noisy hand mask sequence, \moniker{} generates detailed hands including individual fingers, see \cref{fig:noisy}. Though not always consistent, \moniker{} is capable of depicting post-interaction dynamics, such as rolling or sliding objects. 

\vspace{0.25em}
\qheading{Qualitative analysis}
We present diverse qualitative results generated by \moniker{} in \cref{fig:results_diverse}. Row 1 shows how \moniker{} generates the articulated motion of an object. Row 2 showcases pouring water into a glass; note how the water level increases over time. Row 3 demonstrates an object being dropped, moving out of frame when falling, and rolling back in frame once hitting the floor, featuring realistic motion blur synthesis. Rows 4 and 5 illustrate how \moniker{} handles squeezing interactions—the rubber and the spring are compressed and restored accordingly. Row 6 demonstrates an understanding of physical size and distance to the camera, as the phone moves closer to the viewer. These results highlight the complexity that \moniker{} can handle, implying its generalization ability and future potential as an implicit, yet generalized physical simulator and renderer.
\section{Conclusion}
\label{sec:conclusion}

We introduced \moniker{}, a framework that generates videos of interactive dynamics using large video generation models as implicit physics simulators. By incorporating an interactive control mechanism, \moniker{} produces plausible, temporally consistent videos of object interactions—including complex human-object interactions—while generalizing to unseen objects. Our evaluations demonstrate that \moniker{} effectively captures continuous motion and subsequent dynamics, outperforming baselines that focus on single future states. This work highlights the potential of using generative video models as physics simulators without explicit reconstruction, opening new avenues for future research.
 \newpage
{\small\noindent\textbf{Acknowledgements:}
We thank Benjamin Pellkofer for IT support, Peter Kulits for discussions, Tomasz Niewiadomski for coding support, and Allan D.~Jepson, Artur Grigorev, Soubhik Sanyal, Omid Taheri \& Angjoo Kanazawa for feedback. DT is supported by the ERC Starting Grant (project STRIPES, 101165317).}

{\small\noindent\textbf{Disclosure:}
\href{https://files.is.tue.mpg.de/black/CoI\_CVPR\_2025.txt}{https://files.is.tue.mpg.de/black/CoI\_CVPR\_2025.txt}
DT has received a research gift fund from Google.}

{
    \small
    \balance
    \bibliographystyle{config/ieeenat_fullname}
    \bibliography{main}
}

\renewcommand*{\thesection}{\Alph{section}}
\newcommand{\multiref}[2]{\cref{#1}--\ref{#2}}
\renewcommand{\thetable}{S\arabic{table}}
\renewcommand{\thefigure}{S\arabic{figure}}
\setcounter{table}{0}
\setcounter{figure}{0}

\clearpage
\maketitlesupplementary
\appendix

\begin{table*}[b]
    \centering
    \resizebox{\linewidth}{!}{
        \begin{tabular}{@{}lccccccccccccccc@{}}

\toprule

\multirow{2}{*}{Control} &
\multirow{2}{*}{Occlusion} &
\multicolumn{2}{c}{\SSIM $\uparrow$} & &
\multicolumn{2}{c}{\PSNR $\uparrow$} & &
\multicolumn{2}{c}{\LPIPS $\downarrow$} & &
\multicolumn{2}{c}{\FVD $\downarrow$} & &
\multicolumn{2}{c}{\MF $\uparrow$~\cite{yatim2024space}} \\

& &
\squareres & 
\videores & &
\squareres & 
\videores & &
\squareres & 
\videores & &
\squareres & 
\videores & &
\squareres & 
\videores \\

\midrule

Mask &
\xmark &
\textbf{0.829} &
\underline{0.847} & &
24.08 &
24.75 & &
0.123 &
\underline{0.121} & &
\underline{39.94} &
41.99 & &
0.666 &
0.670 \\

Joints &
\xmark &
0.827 &
0.846 & &
24.00 &
24.72 & &
0.124 &
0.122 & &
40.02 &
\underline{41.17} & &
\underline{0.673} &
\underline{0.676} \\

Mesh &
\xmark &
\underline{0.828} &
\underline{0.847} & &
\underline{24.14} &
\underline{24.83} & &
\underline{0.122} &
\underline{0.121} & &
41.99 &
42.26 & &
0.663 &
0.665 \\
    
\midrule

Mask &
\cmark &
\textbf{0.829} &
\underline{0.847} & &
\textbf{24.15} &
24.79 & &
\underline{0.122} &
\underline{0.121} & &
\textbf{37.64} &
41.18 & &
\textbf{0.675} &
0.672 \\

Joints &
\cmark &
0.827 &
0.846 & &
24.05 &
24.69 & &
0.124 &
0.122 & &
44.07 &
41.41 & &
0.665 &
\underline{0.676} \\

Mesh &
\cmark &
\textbf{0.829} &
\textbf{0.848} & &
\textbf{24.15} &
\textbf{24.86} & &
\textbf{0.121} &
\textbf{0.119} & &
40.11 &
\textbf{38.83} & &
\textbf{0.675} &
\textbf{0.680} \\

\bottomrule

\end{tabular}
    }
    \caption{
        \qheading{Ablation on control signal}
        We train and evaluate \moniker{} on \DexYCB~\cite{chao2021dexYCB}. We ablate: type of control signal (mask, joints, and a colored mesh rendering), presence of object occlusions in the control signal, and two image resolutions (\squareres \& \videores).
    } \label{tab:2_dexycb}
\end{table*}

\section{Ablation study}
\label{sec:ablation}

We use binary hand masks as our control signal due to their widespread availability via methods such as \GDINO~\cite{liu2023grounding} and \SAM~\cite{ravi2024sam}. However, other control signals—such as skeletons or meshes—might provide richer controllability, since they encode pseudo-\ThreeD information and fine-grained correspondences across frames.

\begin{figure}[H]
    \centering
    \includegraphics[width=\columnwidth]{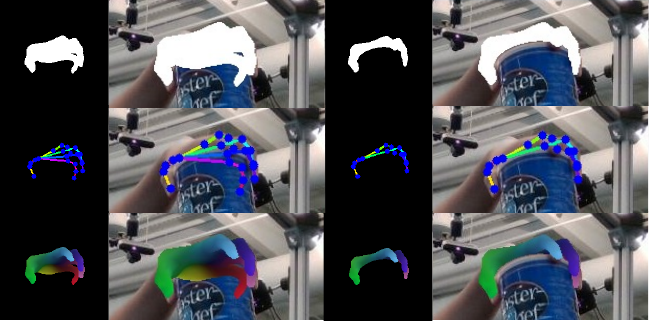}
    \caption{
        \qheading{Evaluated control signals}
        From top to bottom: binary mask, joints in the style of \OPENPOSE~\cite{cao2017realtime}, and colored mesh~\cite{mueller2019real}. Left: w/o object occlusions. Right: w/ object occlusions.
    } \label{fig:control-signal}
\end{figure}

We evaluate the impact of our control signal on performance using the \DexYCB dataset~\cite{chao2021dexYCB}. It provides 8,000 videos of hands grasping an object, along with ground-truth \ThreeD hand/object poses/meshes. \DEXYCB uses the parametric human hand model \MANO~\cite{romero2022embodied}, rendered here as (i) a binary mask (similar to our \SSV control signal), (ii) joints similar to \OPENPOSE~\cite{cao2017realtime}, and (iii) a colormap based on~\cite{mueller2019real}, see \cref{fig:control-signal}. Additionally, when generating hand masks for \SSV with \SAM, hand-held objects provide an object contour in the hand mask, inadvertently providing \moniker{} with information on the trajectory and shape of the object (a limitation we share with our baseline \CosHand~\cite{sudhakar2024coshand}). To evaluate its impact, we train separate \moniker{} versions on \DEXYCB, where we render the control signal either with or without the contour of hand-held objects, see \cref{fig:control-signal}.

We present the ablation results in \cref{tab:2_dexycb}, which indicate that both the type of control signal and contour-leaking effect have minimal impact on image quality, spatio-temporal dynamics, and motion fidelity. These findings softly hint that maintaining hand consistency and driving object interactions does not heavily depend on detailed control signals, rather it does on the video generation model's implicit understanding of interactive dynamics. This highlights the potential of using simple, readily available control signals to generate high-quality video outputs.

\section{State comparison}
\label{sec:state_comparison}

For completeness, we also compare against \CosHand~\cite{sudhakar2024coshand} for the second frame of each video and compare these results with the baselines reported in~\cite{sudhakar2024coshand}, see \cref{tab:results-ssv2-frame}. For \MCVD, \UCG, \IPP, \TCG, and \CosHand (the first five rows), we adopt the results reported in~\cite{sudhakar2024coshand} without retraining. Since \CosHand does not provide, nor specify, an exact validation split, these numbers are not directly comparable. We also run \CosHand on our own validation set.

\begin{table}[hbt!]

    \centering
    
    \resizebox{\linewidth}{!}{
    \begin{tabular}{@{}lccccc@{}}

\toprule

Method &
\SSIM $\uparrow$ & &
\PSNR $\uparrow$ & &
\LPIPS $\downarrow$ \\
    
\midrule

\MCVD~\cite{voleti2024sv3d, sudhakar2024coshand} &
0.231 & &
8.75 & &
0.307 \\

\UCG~\cite{rombach2022high, sudhakar2024coshand} &
0.340 & &
12.08 & &
0.124 \\

\IPP~\cite{brooks2023instructpix2pix, sudhakar2024coshand} &
0.289 & &
9.53 & &
0.296 \\

\TCG~\cite{rombach2022high, sudhakar2024coshand} &
0.234 & &
9.05 & &
0.221 \\

\CosHand~\cite{sudhakar2024coshand} &
0.414 & &
13.72 & &
\textbf{0.116} \\
    
\midrule

\CosHand~\cite{sudhakar2024coshand} (our val-set) &
0.698 & &
20.55 & &
0.194 \\

Ours (256$\times$256) &
\underline{0.785} & &
\underline{23.93} & &
0.127 \\

Ours (256$\times$384) &
\textbf{0.796} & &
\textbf{24.37} & &
\underline{0.122} \\

\bottomrule
    
\end{tabular}}
    \caption{
        \qheading{State comparison on the \SSV dataset}
        We compare \moniker with \CosHand and other static baseline methods for generating a single future frame. We report results for \moniker at two resolutions: 256×256 (matching \CosHand) and 256×384 (matching \SVD's prior).
    } \label{tab:results-ssv2-frame}
    
\end{table}

\begin{figure*}[b]
    \centering
    \includegraphics[width=0.95\textwidth]{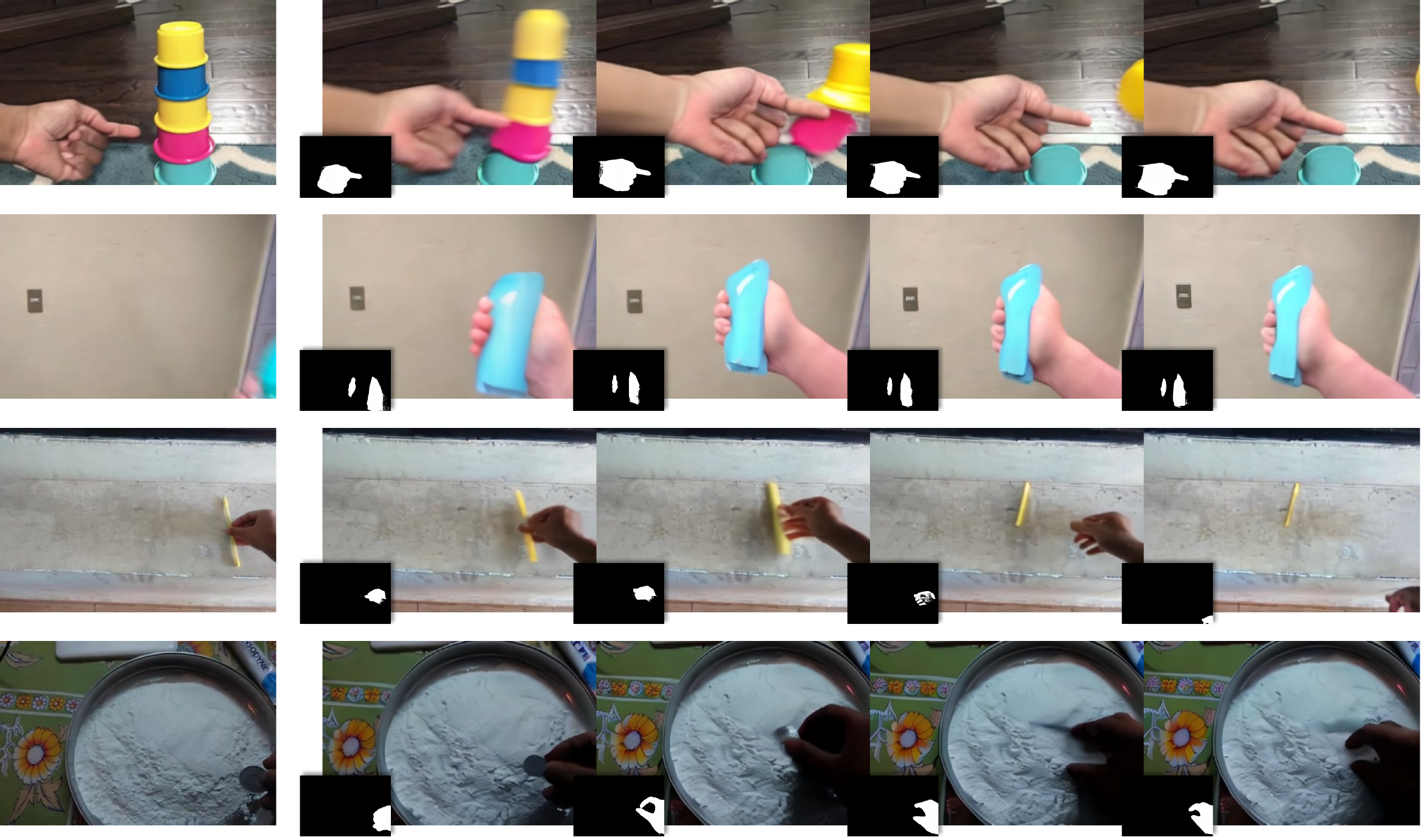}
    \caption{
        \qheading{Limitations of \moniker{}}
        We show challenging scenarios in which \moniker{} underperforms, such as (from top to bottom): object consistency in highly dynamic scenarios, no object in the first frame, depth ambiguity, and burying an object. \faSearch~\textbf{Zoom in} for details.
    } \label{fig:limitations}
\end{figure*}

\section{Limitations}
\moniker{} performs best on translations relative to the camera; dropping objects, moving objects toward or away from the camera, and picking up objects. \moniker{} struggles with complex non-translational interactions (\eg throwing one object at another, burying an object, or poking a tower of stacked objects). It also underperforms in scenarios where depth is ambiguous in the input image, see \cref{fig:limitations}.

\begin{table*}[b]
    \begin{subtable}{0.5\linewidth}
    
        \centering
        \captionsetup{width=0.9\linewidth}

        \resizebox{0.9\columnwidth}{!}{
            \begin{tabular}{
p{11cm}
p{1cm}
p{2cm}
}

\toprule

Label &
Count &
MF $\uparrow$ (avg.) \\
    
\midrule

Moving something down &
182 &
0.86 \\

Pulling something from right to left &
57 &
0.84 \\

Moving something up &
197 &
0.82 \\

Pulling something from left to right &
83 &
0.82 \\

Holding something over something &
165 &
0.80 \\

Holding something &
103 &
0.80 \\

Moving something across a surface without it falling down &
26 &
0.79 \\

Pushing something from left to right &
123 &
0.79 \\

Holding something in front of something &
138 &
0.78 \\

Pushing something from right to left &
122 &
0.77 \\

Putting something on a surface & 85 &
0.77 \\

Moving something across a surface until it falls down &
28 &
0.77 \\

Lifting something with something on it &
369 &
0.77 \\

Squeezing something &
216 &
0.77 \\

Lifting something up completely without letting it drop down &
66 &
0.75 \\

Throwing something in the air and letting it fall &
6 &
0.75 \\

Moving something closer to something &
105 &
0.75 \\

Holding something next to something &
135 &
0.75 \\

Putting something that can't roll onto a slanted surface, so it stays where it is &
15 &
0.75 \\

Trying to bend something unbendable so nothing happens &
74 &
0.74 \\

\bottomrule

\end{tabular}}
        \caption{
            \qheading{Top 20 categories}
            Contains many translation dynamics with respect to the camera, such as moving something up or from left to right.
        } \label{tab:best}
    \end{subtable}%
    \begin{subtable}{0.5\linewidth}
    
        \centering
        \captionsetup{width=0.9\linewidth}
        
        \resizebox{0.9\columnwidth}{!}{
            \begin{tabular}{
p{11cm}
p{1cm}
p{2cm}
}

\toprule

Label &
Count &
MF $\uparrow$ (avg.) \\

\midrule

Spinning something so it continues spinning &
51 &
0.47 \\

Poking something so that it falls over &
42 &
0.46 \\

Pulling something out of something &
33 &
0.46 \\

Folding something &
187 &
0.46 \\

Poking something so it slightly moves &
71 &
0.45 \\

Spinning something that quickly stops spinning &
47 &
0.45 \\

Taking something out of something &
66 &
0.45 \\

Unfolding something &
122 &
0.44 \\

Putting something, something, and something on the table &
60 &
0.44 \\

Piling something up &
27 &
0.43 \\

Something being deflected from something &
10 &
0.41 \\

Poking something so lightly that it doesn't or almost doesn't move &
83 &
0.41 \\

Burying something in something &
4 &
0.41 \\

Showing something next to something &
19 &
0.40 \\

Pushing something so it spins &
23 &
0.39 \\

Poking something so that it spins around &
7 &
0.39 \\

Putting number of something onto something &
5 &
0.37 \\

Poking a stack of something so the stack collapses &
8 &
0.34 \\

Showing something on top of something &
14 &
0.34 \\

Wiping something off of something &
9 &
0.32 \\

\bottomrule
    
\end{tabular}}
        \caption{
            \qheading{Bottom 20 categories}
            Contains very complex dynamics such as spinning, burying, or showing an object from behind something.
        } \label{tab:worst}
    \end{subtable}
    \caption{
        \qheading{Motion fidelity for different action classes on the \SOMETHING dataset}
        The table shows the top and bottom 20 categories, together with the number of samples for that category in the validation set.
    } \label{tab:best_to_worst}
    
\end{table*}

We report the 20 video classes on which \moniker{} \videores performs best and worst in terms of motion fidelity, alongside the number of videos for each class in the validation set and the average motion fidelity score for that class, see  \cref{tab:best_to_worst}. We generally notice that \moniker{} performs less effectively on underrepresented classes within the dataset, while at the same time, many of these underrepresented classes involve complex dynamics, such as spinning, burying, or folding objects.

\begin{figure*}
    \centering
    \includegraphics[width=1.0\textwidth]{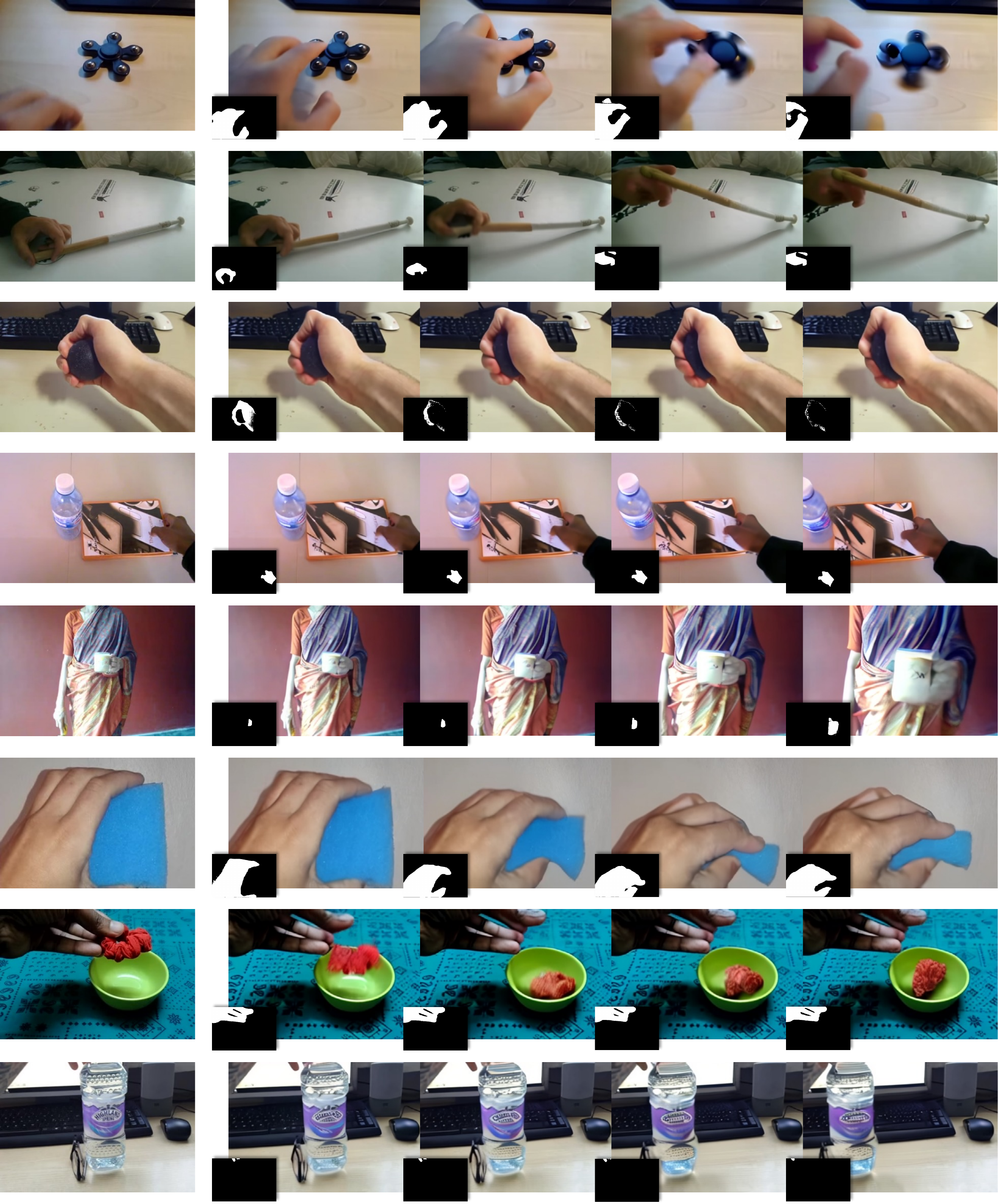}
    \caption{
        \qheading{Additional qualitative results on \SSV}
        We show multiple challenging examples, such as (from top to bottom): spinning a fidget spinner, tilting a sleek ridged object, squeezing a ball despite receiving an incomplete control signal, hand object-object interaction, zooming in, squeezing a sponge, dropping a hairband, or hand object-object interaction despite receiving a sparse control signal.
    } \label{fig:supplementary-results}
\end{figure*}

\section{Additional Results}
We show additional qualitative examples in~\cref{fig:supplementary-results}.

\section{Pretending class}
\label{sec:pretending}

Similar to our baseline \CosHand~\cite{sudhakar2024coshand}, we removed the ``pretending'' class from \SSV for training and validation, to avoid introducing ambiguous training signals. To compare its performance on this class to our results in \cref{tab:results-ssv2-video}, we run \moniker{} on the ``pretending'' class in the validation split (828 samples for \videores and 1156 for \squareres), see \cref{tab:pretending}. While the generations stay consistent in terms of image quality, we notice that motion fidelity is lower. Unfortunately, since \FID, \KID, \FVD, and \KVD compare distributions and are heavily dependent on the number of data samples, we cannot directly compare these metrics to those reported in \cref{tab:results-ssv2-video}.

\begin{table*}[b]

    \centering
    \resizebox{\linewidth}{!}{
        \begin{tabular}{@{}lccccccccccccccc@{}}

\toprule

Method &
\SSIM $\uparrow$ & &
\PSNR $\uparrow$ & &
\LPIPS $\downarrow$ & &
\FID$\downarrow$ & &
\KID$\downarrow$ & &
\FVD$\downarrow$ & &
\KVD$\downarrow$ & &
\MF $\uparrow$~\cite{yatim2024space} \\

\midrule

\seer~\cite{gu2023seer} &
0.357 & &
9.42 & &
0.657 & &
74.86 & &
0.060 & &
640.06 & &
147.07 & &
\NA \\

\dynamicrafter~\cite{xing2024dynamicrafter} $^\dagger$ &
\NA & &
\NA & &
\NA & &
34.96 & &
0.016 & &
314.05 & &
34.22 & &
\NA \\

\mbox{\CosHand-Independent}~\cite{sudhakar2024coshand} &
0.620 & &
16.79 & &
0.310 & &
\textbf{9.85} & &
\textbf{0.003} & &
123.59 & &
15.97 & &
0.396 \\

\mbox{\CosHand-Autoregressive}~\cite{sudhakar2024coshand} &
0.534 & &
14.80 & &
0.410 & &
24.10 & &
0.012 & &
139.07 & &
11.18 & &
0.512 \\

\midrule
    
Ours \squareres &
\underline{0.666} & &
\underline{18.52} & &
\underline{0.256} & &
\underline{14.29} & &
\underline{0.004} & &
\textbf{49.02} & &
\underline{-0.131} & &
\underline{0.573} \\

Ours \videores &
\textbf{0.683} & &
\textbf{18.99} & &
\textbf{0.249} & &
17.39 & &
\underline{0.004} & &
\underline{64.18} & &
\textbf{-0.450} & &
\textbf{0.572} \\

\bottomrule

\end{tabular}
    }
        
    \caption{
        \qheading{Quantitative comparison on the ``pretending'' class of \SSV}
        We compare against \seer~\cite{gu2023seer}, \dynamicrafter~\cite{xing2024dynamicrafter} and two video extensions of our baseline \CosHand~\cite{sudhakar2024coshand}. Methods denoted with $\dagger$ do not use \SSV as their training dataset.
    } \label{tab:pretending}
    
\end{table*}

\end{document}